\documentclass[10pt,onecolumn,twoside]{article}

\usepackage{amsmath,graphicx}
\usepackage{amssymb,bm}
\usepackage{amsmath}
\usepackage{cite}
\usepackage{amsthm}
\usepackage{float}
\usepackage{color}
\usepackage[noend]{algpseudocode}
\usepackage{algorithm,algorithmicx}
\usepackage{fullpage}

\def\defn{\,\triangleq\,}

\def\gbf{{\mathbf{g}}}
\def\ubf{{\mathbf{u}}}

\def\ybf{{\mathbf{y}}}

\def\fbf{{\mathbf{f}}}
\def\pbf{{\mathbf{p}}}
\def\qbf{{\mathbf{q}}}
\def\rbf{{\mathbf{r}}}

\def\drm{{\mathrm{d}}}
\def\Hrm{{\mathrm{H}}}
\def\xbm{{\bm{x}}}

\def\fbfhat{{\widehat{\mathbf{f}}}}
\def\ubfhat{{\widehat{\mathbf{u}}}}

\def\Dbf{{\mathbf{D}}}

\def\Abf{{\mathbf{A}}}
\def\Ibf{{\mathbf{I}}}

\newcommand{\diag}[1]{\mathrm{diag}(#1)}

\def\Rcal{\mathcal{R}}
\def\Dcal{\mathcal{D}}

\def\R{\mathbb{R}}
\def\C{\mathbb{C}}

\def\argmin{\mathop{\mathrm{arg\,min}}}

\def\grad{\nabla}

\newcommand*\Let[2]{\State #1 $\gets$ #2}

\title{Compressive Imaging with Iterative Forward Models} 

\author{Hsiou-Yuan~Liu%
\thanks{H.-Y.~Liu (email: hyliu@eecs.berkeley.edu) is with 
Department of Electrical Engineering \& Computer Sciences, University of California, Berkeley, CA 94720, USA. This work
was completed while he was with Mitsubishi Electric Research Laboratories (MERL).}
\hspace{0.05em},
Ulugbek~S.~Kamilov%
\thanks{U.~S.~Kamilov (email: kamilov@merl.com), D.~Liu (email: liudh@merl.com), H.~Mansour (email:mansour@merl.com), and P.~T.~Boufounos (email: petrosb@merl.com)
are with Mitsubishi Electric Research Laboratories (MERL), 201 Broadway, Cambridge,
MA 02139, USA.}
\hspace{0.05em}, Dehong~Liu$^\dagger$,
\\Hassan~Mansour$^\dagger$, and Petros~T.~Boufounos$^\dagger$}

\begin{document}

\maketitle


\begin{abstract}
We propose a new compressive imaging method for reconstructing 2D or 3D objects from their scattered wave-field measurements. Our method relies on a novel, nonlinear measurement model that can account for the multiple scattering phenomenon, which makes the method preferable in applications where linear measurement models are inaccurate. We construct the measurement model by expanding the scattered wave-field with an accelerated-gradient method, which is guaranteed to converge and is suitable for large-scale problems. We provide explicit formulas for computing the gradient of our measurement model with respect to the unknown image, which enables image formation with a sparsity-driven numerical optimization algorithm. We validate the method both analytically and with numerical simulations.
\end{abstract}


\section{Introduction}
\label{Sec:Introduction}

Some of the most difficult, yet important, problems in computational sensing involve imaging objects that are hidden behind an opaque medium. For example, identifying a tumor inside a human body in medical diagnosis, detecting defects within a structure in industrial testing, or visualizing the shape of a multicellular organism in biology are all instances of this fundamental problem of subsurface imaging.
The most commonly used approach in such applications is based on probing the medium with a controlled incident wave of a specific frequency range that can penetrate the medium, and then to rely on the physics of wave scattering to infer or visualize the spatial distribution of the refractive index within the medium.

This problem of inferring the refractive index distribution from the scattered wave-field is known as inverse scattering. It is often formulated as a large-scale optimization that relies on models for describing both the physical (forward or measurement model) and signal-related (regularization, prior constraints) aspects of the problem. Traditional approaches to inverse scattering are based on linear measurement models that can be obtained by assuming a straight-ray propagation of waves~\cite{Kak.Slaney1988}, or by adopting more refined scattering models based on the first Born~\cite{Born.Wolf1999} or Rytov approximations~\cite{Devaney1981}. Recent works have demonstrated impressive imaging capability of the optimization-based approaches that also incorporate prior constraints on the solution~\cite{Choi.etal2007, Sung.etal2009, Kim.etal2014}. In particular, dramatic improvements were obtained by relying on sparsity-promoting regularization~\cite{Bronstein.etal2002, Sung.Dasari2011, Lim.etal2015}, which is an essential component of compressive sensing~\cite{Candes.etal2006, Donoho2006}. The basic motivation is that many natural objects are inherently sparse in some transform domain and can be reconstructed with high accuracy even with a small amount of measured data.

Linear measurement models, though simple and efficient, are only accurate for weakly scattering objects. This limits their applicability for imaging larger objects and/or those with large refractive index contrasts. Recent experimental results also indicate that the resolution and quality of the reconstructed image is improved when nonlinear measurement models are used~\cite{Simonetti2006, Simonetti.etal2008, Maire.etal2009, Tian.Waller2015, Kamilov.etal2015, Kamilov.etal2016, Kamilov.etal2016a, Zhang.etal2016}. In particular, nonlinear models can account for multiple scattering and thus provide a more accurate interpretation of the measured data.

We propose a new method for reconstructing the refractive index from measurements of the scattered wave-field. This method combines our nonlinear forward model with an edge-preserving total variation (TV) regularizer~\cite{Rudin.etal1992} and forms images by solving a large-scale optimization problem. Our measurement model---called series expansion with accelerated gradient descent on Lippmann-Schwinger equation (SEAGLE)---is based on formulating wave-scattering as a smooth optimization subproblem and using Nesterov's fast gradient method~\cite{Nesterov1983} to iteratively approximate the scattered wave. The key advantage of SEAGLE is its guaranteed convergence, even for objects with large refractive index contrasts. We provide explicit formulas for computing the gradient of our measurement model with respect to the refractive index, which enables large-scale 2D and 3D imaging using fast iterative shrinkage/thresholding algorithm (FISTA)~\cite{Beck.Teboulle2009}. We validate our forward model and reconstruction method analytically and with numerical simulations.

\section{Main Contribution}
\label{Sec:mainc}
Our approach expands the scattered wave field with the iterates of Nesterov's accelerated gradient descent and efficiently computes the derivative of the field with respect to the object. The inverse problem is formulated as a TV-regularized data fitting with complex scattered-wave measurements. 

\subsection{Problem formulation}\label{sec:pb-formulate}

\begin{figure}
\centering
\includegraphics[width=0.4\textwidth]{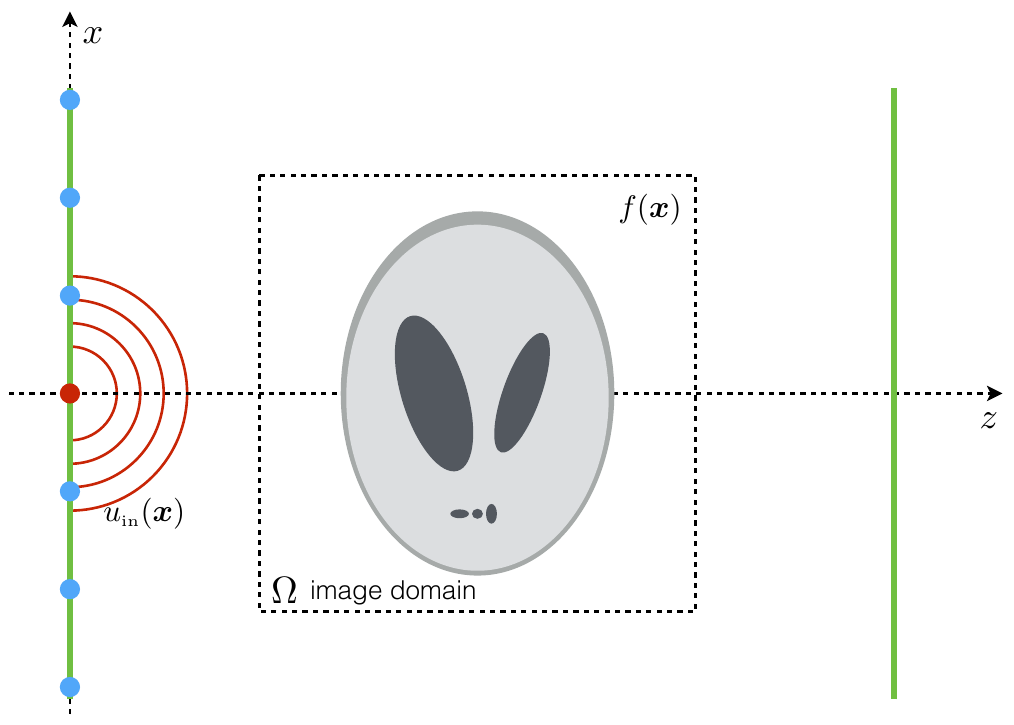}
\caption{\label{fig:scheme} A schematic representation of a scattering scenario. An object with a scattering potential $f(\xbm)$, $\xbm \in \Omega$, is illuminated with an input wave $u_{\text{\tiny in}}$, which interacts with the object and results in the wave-field $u$ measured at two sensor regions (solid lines in the left and right).}
\end{figure}

Consider a setup illustrated in Fig.~\ref{fig:scheme} where the unknown object resides inside the image domain $\Omega \subseteq \R^D$, where ${D \in \{2, 3\}}$ .
We want to recover the refractive index of the unknown object given wave measurements at $M$ point sensors (on the vertical solid lines) and controllable sources (solid circles). Monochromatic light scattering by nonuniform refractive index can be described by the Lippmann-Schwinger equation
\begin{equation}\label{eq:lippmann}
u(\xbm) = u_\mathrm{in}(\xbm) + \int_\Omega g(\xbm-\xbm')f(\xbm')u(\xbm')\drm\xbm'
\end{equation}
where
$u(\xbm)$ is the complex total electric field,
$u_\mathrm{in}(\xbm)$ is the complex incident electric field,
$g(\xbm)$ is the free-space Green's function,
$f(\xbm) \defn k_b^2(1-n(\xbm)^2)$ is the scattering potential, which is assumed to be real and contains the map of refractive index of the object $n(\xbm)$, and $k_b$ is the wavenumber of the background medium.
This integral is only over domain $\Omega$ as $f(\xbm)$ is zero outside of $\Omega$. The free-space Green's function in~\eqref{eq:lippmann} is given by
\begin{equation}\label{eq:greenfunc}
g(\xbm) \defn \left\{\begin{array}{ll}
-\frac{\mathrm{j}}{4}H_0^{(1)}(k_br)                &\text{for }D = 2\\
\displaystyle -\frac{e^{\mathrm{j}k_br}}{4\pi r}    &\text{for }D = 3
\end{array}
\right.,
\end{equation}
where $r=\|\xbm\|_2$ and $H_0^{(1)}$ is the Hankel function of first kind. The Green's function is obtained under the outgoing wave boundary condition, Helmholtz equation $\left(\nabla^2 + k_b^2\right) g(\xbm) = +\delta(\xbm)$, and the time-dependence convention where the physical electric field equals to $\mathfrak{Re}\{u(\xbm)e^{-\mathrm{j}\omega t}\}$.

Eq.~\eqref{eq:lippmann} provides a nonlinear relationship between the wave-field $u$ and the scattering potential $f$, whereas first Born and Rytov approximations provide simplified linearized versions of this relationship. The inverse problem is to find an estimation of $f$ given the measurements of $u$ at point sensors.

\subsection{Algorithmic Expansion of the Wave Model}



For points in the domain $\Omega$, the discretized version of~(\ref{eq:lippmann}) can be expressed as
\begin{equation}\label{eq:lippmann-dis}
\ubf = \ubf_{\text{\tiny in}}+\mathbf{G} \, \diag{\fbf} \,\ubf,
\end{equation}
where the operator $\diag{\cdot}$ forms a diagonal matrix from its argument, $\ubf,\ubf_\mathrm{in}\in \mathbb{C}^N$, and $\fbf\in \R^N$ are the discretized versions of $u$, $u_\mathrm{in}$ and $f$, respectively, and $\mathbf{G}\in\mathbb{C}^{N\times N}$ is the matrix representing the convolution with the Green's function within $\Omega$. Here, $N$ denotes the number of sample points. Note that the field outside $\Omega$ can be evaluated by using a different matrix $\mathbf{\tilde G}$, which corresponds to evaluating~\eqref{eq:lippmann} at sensor points. 

As a forward model, we propose to solve~\eqref{eq:lippmann-dis} for $\ubf$ by applying Nesterov's fast gradient method to the following minimization problem
\begin{equation}\label{eq:lippmann-opt}
\ubfhat(\fbf) \defn \argmin_{\ubf \in \C^N} \left\{\frac12 \left\|\Abf\ubf - \ubf_\mathrm{in}\right\|_2^2\right\},
\end{equation}
where $\Abf \defn \mathbf{I}-\mathbf{G} \, \diag{\fbf}$. The full procedure is summarized in Algorithm~\ref{alg:forward}, which we call SEAGLE. Note the dependence of the solution $\ubfhat$ on the scattering potential $\fbf$ and the fact that it can be interpreted as an expansion of the wave-field with the iterates of the accelerated gradient descent method.

\begin{algorithm}
\caption{Forward model computation \label{alg:forward}}
  \begin{algorithmic}[1]
    \Require{$\ubf_\mathrm{in}$, $\fbf$, $\mathbf{G}$, $\mathbf{\tilde G}$, number of iterations $K$, tolerance $g_{\text{tol}}$ and initial field $\ubf_\mathrm{init}=\ubf_\mathrm{in}$}
        \State $\ubf_{-1} \gets \ubf_\mathrm{init},\ \ubf_0 \gets \ubf_\mathrm{init},\ t_0 \gets 0$
        \For{$k \gets 1 \textrm{ to } K$}
            \State $t_k \gets (1+\sqrt{1+4\smash[b]{t_{k-1}^2}})/{2}$, 
            \State $\mu_k \gets ({t_{k-1}-1})/{t_k}$
            \Let{$\ybf_k$}{$\ubf_{k-1} + \mu_k(\ubf_{k-1}-\ubf_{k-2})$}
            \Let{$\mathbf{v}$}{$\Abf^\Hrm(\Abf\ybf_k-\ubf_\mathrm{in})$} \label{ln:gradient}\Comment{gradient at $\ybf_k$}
            \If{$\| \mathbf{v} \|_2 < g_{\text{ tol}}$} $K \gets k$, break for loop
            \EndIf
            \Let{$\gamma_k$}{$\smash[b]{\| \mathbf{v} \|^2_2 / \| \Abf\mathbf{v} \|^2_2}$} \label{ln:forward-lipschitz}
            \Let{$\ubf_k$}{$\ybf_k-\gamma_k\mathbf{v}$}  \label{ln:update-u}
        \EndFor
        \Let{$\ubfhat$}{$\ubf_\mathrm{in}+\mathbf{\tilde G}\diag{\fbf}\ubf_K$} \label{ln:last-scat}
        \State \Return{the predicted field at sensors $\ubfhat$, as well as $\ubf_K$, $\{\gamma_k\}$, $\{\ybf_k\}$,  and $\{\mu_k\}$}
  \end{algorithmic}
\end{algorithm}

\subsection{Inverse problem}

We formulate the inverse problem as the minimization
\begin{equation}\label{eq:inverse}
\fbfhat = \argmin_{\fbf \in\mathcal{F}} \left\{\Dcal(\fbf) + \tau \Rcal(\fbf) \right\},
\end{equation}
where
\begin{subequations}
\begin{align}
&\Dcal(\fbf) \defn \frac{1}{2} \left\| \ubfhat(\fbf) - \mathbf{m} \right\|_2^2 \\
&\Rcal(\fbf) \triangleq \sum_{n=1}^N \sqrt{\sum_{d = 1}^D ([\Dbf_d \fbf]_n)^2}.
\end{align}
\end{subequations}
Here, $\Dcal$ is the quadratic data-fidelity term that measures the discrepancy between the measured data $\mathbf{m}\in \mathbb{C}^M$ and the output of SEAGLE forward computation $\ubfhat$. The functional $\Rcal$ is $D$-dimensional isotropic TV regularizer, where $\Dbf_d$ is the discrete gradient operators along the axis $d$.  The regularization parameter $\tau > 0$ controls the strength of regularization, while the set $\mathcal{F} \subseteq \mathbb{R}^N$ is used for enforcing additional physical constraints on $\fbf$ such as, for example, non-negativity.

The two key steps of FISTA for solving the optimization problem~(\ref{eq:inverse}) are computing the $\nabla\Dcal$ and evaluating the proximal operator
\begin{equation}\label{eq:proximal}
\mathrm{prox}_{\alpha\Rcal}(\gbf) \triangleq  \argmin_{f\in \mathcal{F}} \left\{ \frac12\|\fbf-\gbf\|_2^2 + \alpha\Rcal(\fbf)\right\}
\end{equation}
for some $\alpha > 0$ and $\gbf \in \R^N$~\cite{Beck.Teboulle2009}. The proximal operator corresponding to TV can be efficiently computed~\cite{Beck.Teboulle2009a,Kamilov2016}. 

The iterative structure of the SEAGLE forward computation allows for an efficient computation of the gradient, which can be expressed as 
\begin{equation}\label{eq:bp-grad}
\grad \Dcal(\fbf) = \mathfrak{Re}\left\{ \left[\frac{\partial\ubfhat}{\partial\fbf}\right]^\Hrm \left(\ubfhat(\fbf) - \mathbf{m} \right) \right\},
\end{equation}
with $(\frac{\partial \ubfhat}{\partial\fbf})_{ij}=\frac{\partial \widehat u_i}{\partial f_j}$. By differentiating lines \ref{ln:last-scat}, \ref{ln:update-u} and \ref{ln:gradient} in Algorithm~\ref{alg:forward} with respect to $\fbf$ and keeping $\{\gamma_k\}$ constant, we obtain an efficient error-back propagation rule for computing~\eqref{eq:bp-grad} summarized in Algorithm~\ref{alg:backward}. Note that the matrices $\mathbf{S}_k,\mathbf{T}_k$ in Algorithm~\ref{alg:backward} are implemented as operators on $\qbf_k$ without the need for an explicit storage in memory. The remarkable aspect of Algorithm~\ref{alg:backward} is that it explicitly provides the gradient $\nabla \Dcal$ that can be used for FISTA-based minimization of~\eqref{eq:inverse}. While the optimization problem~\eqref{eq:inverse} is generally non-convex, we did not observe any practical convergence issues in our simulations (see Section~\eqref{Sec:numerical}).

\begin{algorithm}
\caption{Gradient computation\label{alg:backward}}
  \begin{algorithmic}[1]
    \Require{$\mathbf{m}$, $\ubfhat$, $\fbf$, $\ubf_\mathrm{in}$, $\mathbf{G}$, $\mathbf{\tilde G}$, $\{\gamma_k\}$, $\ubf_K$, $\{\ybf_k\}$, $\{\mu_k\}$}
        \State $\pbf_K \gets \mathbf{0}$, 
        \Let{$\qbf_K$}{$\diag{\fbf}^\Hrm\mathbf{\tilde G}^\Hrm(\ubfhat-\mathbf{m})$}
        \Let{$\rbf_K$}{$\mathrm{diag}(\ubf_K)^\Hrm\mathbf{\tilde G}^\Hrm(\ubfhat-\mathbf{m})$}
        \For{$k \gets K \textrm{ to } 1$}
        	\State {$\mathbf{S}_k$} $\triangleq$ {$\Ibf-{\gamma_k}\Abf^\Hrm\Abf$}
        	\State {$\mathbf{T}_k$} $\triangleq$ {$\mathrm{diag}(\mathbf{G}^\Hrm(\Abf\ybf_k-\ubf_\textrm{in}))^\Hrm+\mathrm{diag}(\ybf_k)^\Hrm\mathbf{G}^\Hrm\Abf$}
            \Let{$\pbf_{k-1}$}{$-\mu_k\mathbf{S}_k \qbf_k$}
            \Let{$\qbf_{k-1}$}{$\pbf_k+(1+\mu_k)\mathbf{S}_k\qbf_k$}
            \Let{$\rbf_{k-1}$}{$\rbf_k+{\gamma_k}\mathbf{T}_k\qbf_k$}
        \EndFor
        \State \Return{$\grad \Dcal(\fbf) =\mathfrak{Re}(\rbf_0)$ the gradient in~(\ref{eq:bp-grad})}
  \end{algorithmic}
\end{algorithm}

\subsection{Network interpretation}
\begin{figure}
\centering
\includegraphics[width=0.45\textwidth]{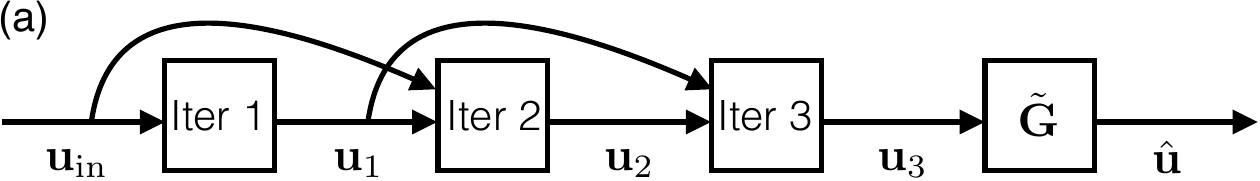}\\[5pt]
\includegraphics[width=0.45\textwidth]{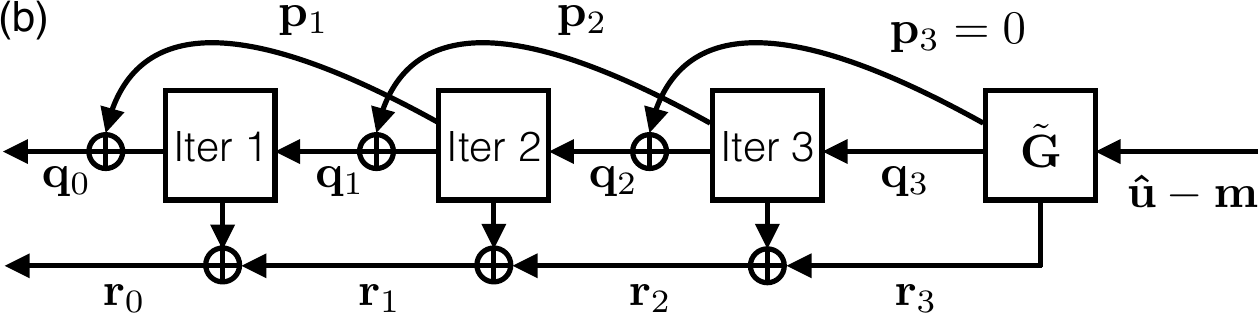}
\caption{ \label{fig:network}The network interpretation of SEAGLE ($K=3$). (a)~forward computation. (b)~gradient propagation}
\end{figure}

Figure~\ref{fig:network} graphically illustrates Algorithms~\ref{alg:forward} and \ref{alg:backward} as feedforward networks. Each square module represents the operation in each iteration of SEAGLE, and the edges represent the vectors as message carriers. A nice feature of SEAGLE is that it can easily incorporate additional modules for modeling other physical phenomena. For example, we can prepend a module representing an initialization with Rytov approximation in front of the module Iter 1. Then, in the back propagation, $\qbf_0$ and $\rbf_0$ are fed into the Rytov module in which the Rytov inverse step is applied. The flexibility of SEAGLE will be explored in future works for speeding up the computation or for dealing with other measurement scenarios where only intensity of the wave-field is preserved.

\section{Numerical Results}
\label{Sec:numerical}

\subsection{Analytic validation of the forward model}

\begin{figure}[t]
\centering
\includegraphics[width=0.45\textwidth]{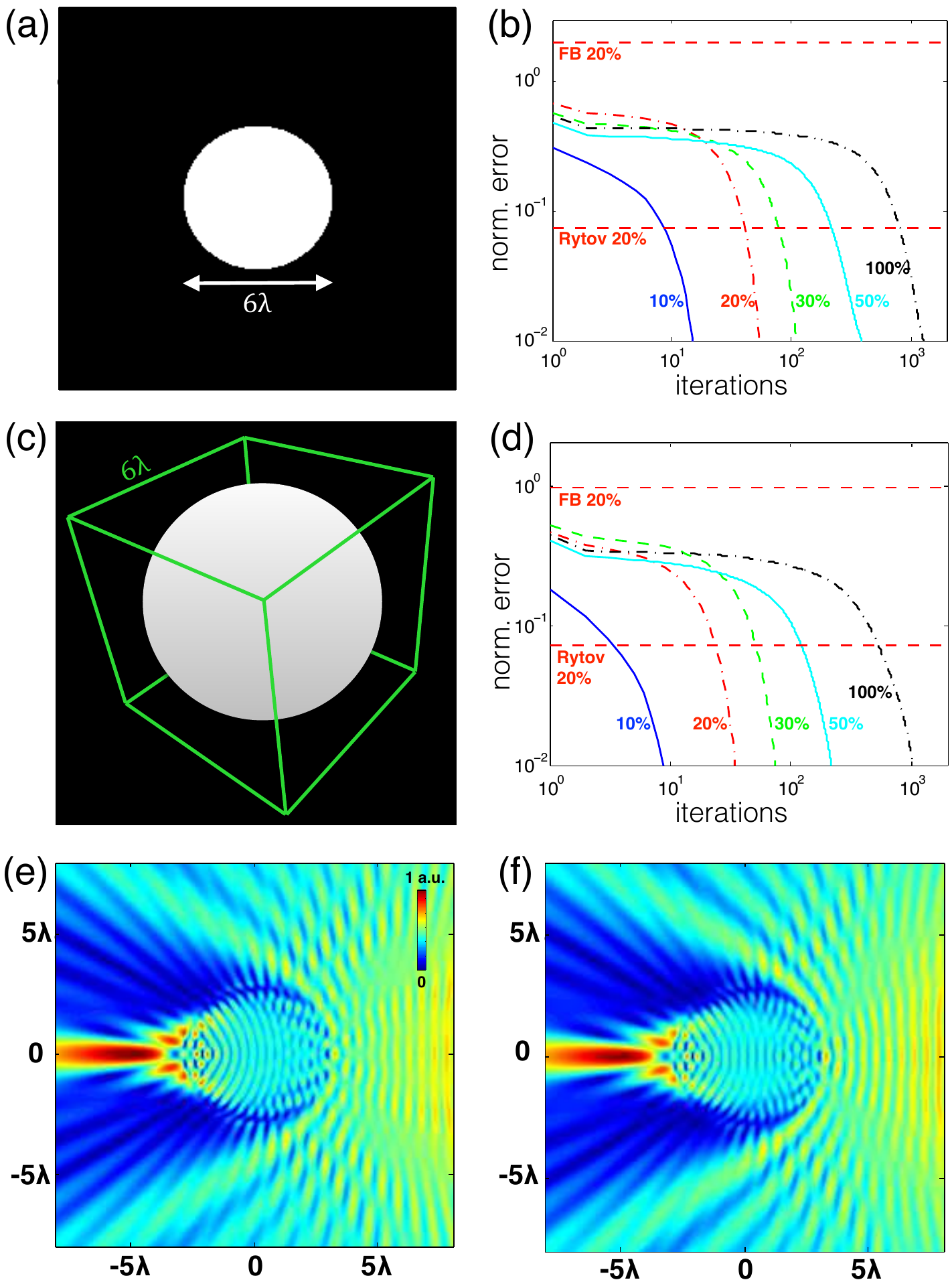}
\caption{\label{fig:compAnalytic} Analytical validation of the proposed measurement model:
(a) A cylinder with a diameter of 6 wavelengths.
(b) Normalized errors for scattering from cylinders of varying contrast levels.
(c) A sphere with a diameter of 6 wavelength.
(d) Normalized error for scattering from spheres of varying contrast levels.
(e) Analytic field for a cylinder with a contrast level of $100\%$.
(f) Corresponding field computed by our forward model.}
\end{figure}

In order to validate the forward model, we first consider two simple scattering experiments where it is possible to derive analytic forms of the scattered wave-fields: a 2D point source scattered by a cylinder, and a 3D point source scattered by a sphere (see Sections 3.8-3.11 in \cite{jackson1999classical}). In both cases the scatterers have diameters of $6$ wavelengths (Fig.~\ref{fig:compAnalytic}(a) and \ref{fig:compAnalytic}(c)). The wavelength is $74.9$~mm, the grid size is $4.8$~mm ($6$~mm) and there are $250$ points ($128$ points) along each axis in 2D (3D). We define the contrast of an object as $\max(|\fbf|)/k_b^2$. In order to evaluate the performance quantitatively, we plot in Fig.~\ref{fig:compAnalytic}(b) and \ref{fig:compAnalytic}(c) the normalized error, $\|\ubfhat-\ubf_\textrm{truth}\|_2^2 / \|\ubf_\textrm{truth}\|_2^2$, where $\ubf_\textrm{truth}$ is the analytic solution. We additionally provide errors achieved when using the first Born and Rytov approximations at $20 \%$ contrast. In Fig.~\ref{fig:compAnalytic}(e) and \ref{fig:compAnalytic}(f), we provide visual comparison between the analytic solution and the result of our model. 

\subsection{Inverse scattering experiment} 

\begin{figure}[t]
\centering
\includegraphics[width=0.45\textwidth]{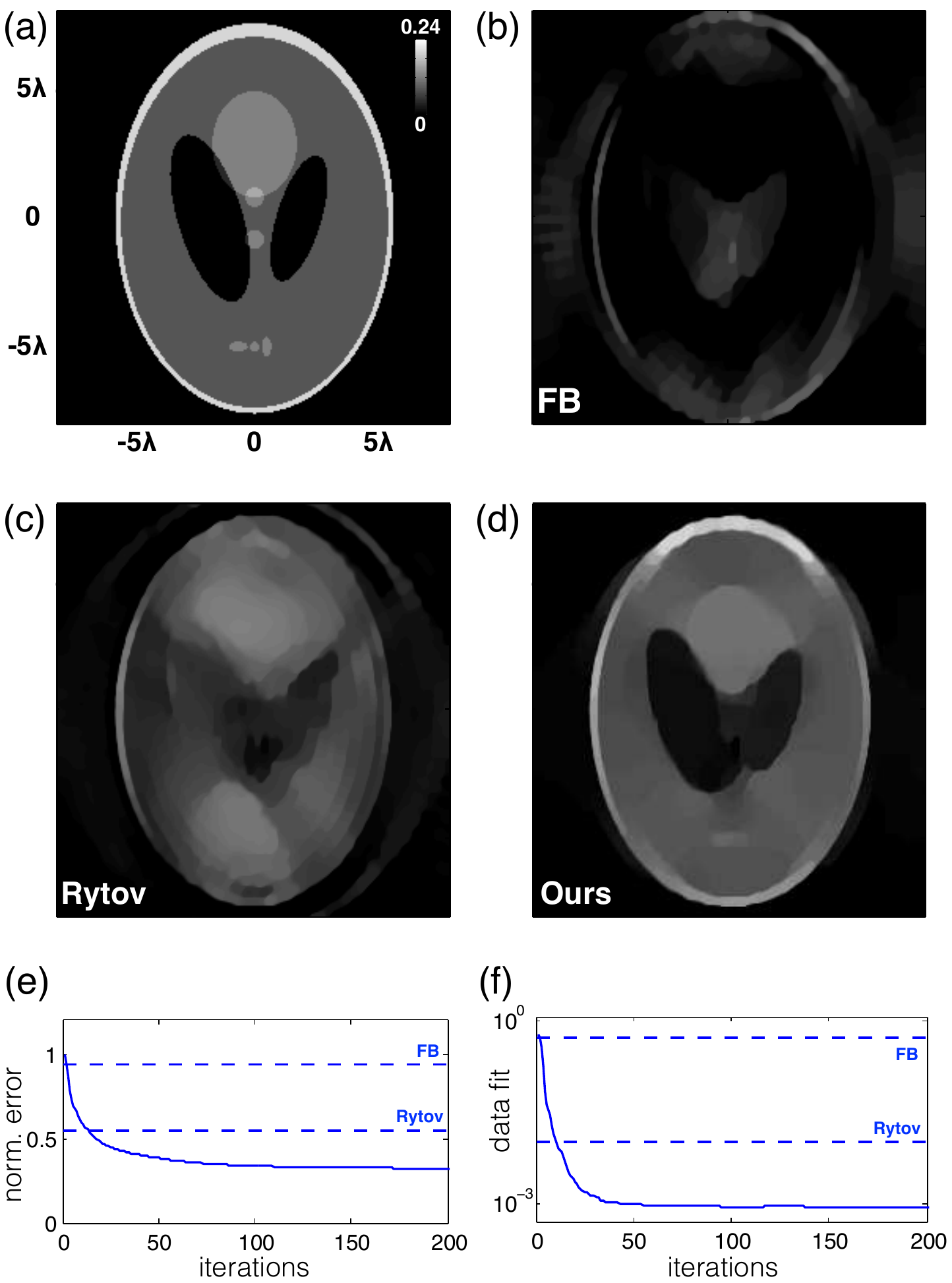}
\caption{ \label{fig:inverse} Reconstruction of the \emph{Shepp-Logan} phantom of $20 \%$ contrast.
(a) True image.
(b)-(d) The reconstructed results with first Born (FB) approximation, Rytov approximation, and our method.
(e) Evolution of the normalized reconstruction error.
(f) Evolution of the normalized data fit $(\| \hat\ubf(\fbf) - \mathbf{m} \|_2^2/\|\mathbf{m}\|_2^2)$}
\end{figure}

We next use the proposed technique for reconstructing the Shepp-Logan phantom in the ill-posed and strongly scattering regime. 
Specifically, we consider the setup in Fig.~\ref{fig:scheme} where the scattered wave measurements are generated by a high-fidelity finite-difference time-domain (FDTD)~\cite{Taflove.Hagness2005} simulator. The object is of size $84.9$~cm $\times$ $113$~cm and has a contrast of $20\%$. We put two linear detectors on both sides of the phantom at a distance of $95.9$~cm from the center of the object, and each detector has $169$ sensors placed with in-between spacing of $3.84$~cm. The transmitters are put on a line $48.0$~cm  left to the left detector. They are spaced uniformly in azimuth with respect to the center of the phantom (every $5^\circ$ within $\pm60^\circ$). We setup a $120$~cm $\times$ $120$~cm square area for reconstructing the object, with pixel size $0.479$~cm. The wavelength of the illuminating light is $7.49$~cm.

Fig.~\ref{fig:inverse} summarizes the performances of the proposed method, as well as two baseline methods based on the first Born and Rytov approximations. All three methods are implemented iteratively with isotropic TV regularizer ($\tau=1.5\times10^{-9}\|\mathbf{m}\|^2$). In SEAGLE, we set $K=120$ but the forward algorithm may stop earlier when the objective function~(\ref{eq:lippmann-opt}) is below $5\times10^{-7}\times\|\ubf_\textrm{in}\|_2^2$. It is shown that SEAGLE outperforms first Born and Rytov methods. It can be seen that due to the ill-posed nature of the measurements, the reconstructed images suffer from the missing frequency artifacts~\cite{sheppard1990three}. However, our method is still able to accurately capture most features of the object while the linear methods cannot.

\section{Conclusion}
\label{Sec:conclusion}

Our method is suitable for compressive imaging in the presence of multiple scattering. It can handle both transmission and reflection data, unlike alternative methods based on beam propagation. It is additionally more stable compared to the methods based on iterative Born approximations.


\bibliographystyle{IEEEtran}

\end{document}